\documentclass{rob}%

\usepackage{amssymb}
\usepackage{amsmath}
\usepackage[super]{cite}
\usepackage{algorithm}
\usepackage[noend]{algpseudocode}
\usepackage{xcolor}
\usepackage{tikz}
\usepackage{ulem}
\usepackage[printwatermark]{xwatermark}

\usepackage{graphicx}

\doi{10.1017/xxxx}
\volume{31}
\issue{5}
\pubyear{2013}
\cpyear{2013}
\endpage{7}

\newwatermark[allpages,color=gray!25,angle=45,scale=3,xpos=70,ypos=-180]{Accepted}

\begin{document}

\title[Towards One-Dollar Robots]{Towards One-Dollar Robots: An Integrated Design and Fabrication Strategy for Electromechanical Systems\vspace{1.5em}}

\author{Wenzhong Yan$\dagger$\thanks{Corresponding author. E-mail:
wzyan24@g.ucla.edu}, and Ankur Mehta$\S$}

\affil{$\dagger$Mechanical and Aerospace Engineering Department, UCLA, Los Angeles, California\\
{\S}Electrical and Computer Engineering Department, UCLA, Los Angels, California}

\ADaccepted{MONTH DAY, YEAR. First published online: MONTH DAY, YEAR}

\maketitle
\begin{summary}

To improve the accessibility of robotics, we propose a design and fabrication strategy to build low-cost electromechanical systems for robotic devices. Our method, based on origami-inspired cut-and-fold and E-textiles techniques, aims at minimizing the resources for robot creation. Specifically, we explore techniques to create robots with the resources restricted to single-layer sheets (e.g. polyester film) and conductive sewing threads. To demonstrate our strategy’s feasibility, these techniques are successfully integrated into an electromechanical oscillator (about 0.40 USD), which can generate electrical oscillation under constant-current power and potentially be used as a simple robot controller in lieu of additional external electronics.
\end{summary}


\section{INTRODUCTION}
Origami robots have long been introduced into the robotics community to achieve rapid and inexpensive prototyping\cite{Rus2018}. However, current origami methodologies require significant resources, including both hardware and domain expertise, and thus preclude the further promotion of robotics' accessibility. To address this challenge, a new design and fabrication strategy that is capable of minimizing the resources required to create robotic systems is strongly desired for improving the accessibility of robotics.

Origami-inspired engineering, as a top-down approach, enables unprecedented rapid prototyping and customization of robots by manufacturing structures in a 2D plane and folding them into their final 3D shapes. The power of origami-inspired planar design and fabrication has been found in several applications\cite{kim_origami-inspired_2018,bassik_microassembly_2009,okuzaki_biomorphic_2008}. Nevertheless, the challenge remains for complicated electromechanical robotic systems. Recently, Onal et al.\cite{rus2013} developed a general principle of building worm robots, whose bodies are made of only a flat sheet and whose actuation is realized with NiTi coil actuators placed on the robot's body. Mehta et al.\cite{mehta2014} proposed a method to cogenerate mechanical, electrical, and software designs for printable robots from structural specifications. So far, origami folding is mostly harnessed to form the mechanical subsystems of origami robots, while the electrical subsystems are still constructed with traditional bulky electronic components and wiring. This method is widely adopted thanks to its simplicity and minimal design iterations required. In other words, the origami-inspired method is merely a rapid and inexpensive alternative for conventional manufacturing approaches (e.g. molding), while the barrier in constructing the robots' electrical subsystems and in integrating robot subsystems remains unsolved. Subsequently, some attempts have been made to incorporate the mechanical and electrical subsystems into integrated systems that can be created through the origami-inspired method. Onal et al.\cite{onal2015} devised smart laminates that contain an electronic layer for the control and actuation of the folded mechanism. Felton et al.\cite{Felton644} then proposed more sophisticated multilayer laminates including self-folding hinges that can be controlled by embedded heating elements, resulting in self-folding functional machines. This class of laminates are found very useful in various applications owing to their capabilities of forming functional, complicated 3D mechanisms (e.g. reconfigurable robots\cite{hawkes_programmable_2010}). However, these laminates require expensive materials and elaborated fabrication processes, which seriously limit its accessibility to casual end-users. In summary, current design and fabrication strategies for origami robots either depend on carefully elaborated materials or merely replace mechanical structures with their origami counterparts, yet still requiring conventional electrical components and software. Therefore, these design and fabrication strategies have limitations in maximizing the accessibility of robotic creation under resource constraints.

Recently, conductive thread has been extensively employed in wearable devices\cite{maziz_knitting_2017,lovell_soft_nodate,bashir_conjugated_2013,mostafalu_toolkit_2016,teng_integrating_2018}. These conductive threads could be fabricated into different configurations that function as soft circuits, switches\cite{qiu_curriculum_2013,harnett_flexible_2016}, antennae\cite{zhang_integration_2018,zhang_novel_2019}, actuators\cite{yan2018,Yip2015}, and sensors\cite{sundaram_learning_2019,zhao_low-cost_2016}, etc. Specifically, the low cost and versatility of this conductive thread have drawn our attention thanks to its great potential to be integrated into origami robots to achieve the ultimate goal of minimizing the resources required for robot creation. Nevertheless, few attempts have been made to incorporate these threads into origami robots, leaving the integration a challenge. 

Aiming at minimizing the required resources, we present a new design and fabrication strategy for building electromechanical systems for origami robots. Specifically, in this paper, we explore methods to create robots under the constraint of only using ordinary single-layer sheet materials (e.g. polyester film) and conductive threads. Toward this goal, four major challenges are recognized and addressed to create functional electromechanical systems, resulting in several design and fabrication principles. These enable us to construct complicated 3D origami architectures, form connections between different features, create electrical contacts, and budget thermal flow under the constraint of only using ordinary single-layer sheet materials and conductive threads. To better demonstrate these design and fabrication principles, we propose various examples, which are common building blocks for constructing electromechanical systems within our considerations and thus can be integrated to generate more sophisticated systems. To validate the feasibility of our integrated design and fabrication strategy, we create a foldable electromechanical oscillator by incorporating the aforementioned principles. These techniques may be also incorporated into various other electromechanical systems to generate additional functionalities on top of oscillation for origami-inspired robots. In addition, the potential of building one-dollar robots based on this oscillator is discussed at the end of the paper.

The contributions of this paper are as follows:
\begin{itemize}
  \item the understanding and recognition of essential challenges for robot creation under the constraint of only using ordinary single-layer sheet materials and conductive threads;
  \item a design and fabrication strategy for creating electromechanical systems for origami robots aiming at minimizing required resources; and 
  \item a case study that demonstrates the integrated design and fabrication strategy.
\end{itemize}

The remainder of this paper is organized as follows: the design and fabrication strategy is introduced in Section \ref{sec:design&fab}; the foldable electromechanical oscillator is proposed in Section \ref{sec:example} to demonstrate our method; we end with conclusion and future works in Section \ref{sec:conclusionAndDiscussion}.

\section{Design and Fabrication Methods}
\label{sec:design&fab}
Aiming at minimizing the resources required for making robotic devices, we propose a design and fabrication strategy for creating low-cost 3D electromechanical devices, with the final goal of building one-dollar functional robots and improving the accessibility of robotics. Specifically, we explore the possibility of creating robots under the constraint of only using ordinary single-layer sheet materials (e.g. polyester film) and conductive threads (or SCP actuators\cite{Yip2015}). Four main challenges, out-of-plane folding, structure fastening, electrical contact forming, and thermal budgeting, are identified along our progress. To tackle these challenges, we combine the versatility of 3D folding methods with the simplicity and material availability of the e-textile technology. In other words, we explore new methods to build 3D electromechanical systems using ordinary single-layer sheet materials and conductive threads (or SCP actuators), without involving complicated fabrication processes or expensive sophisticated building materials. In this paper, we use the combination of commercially available, flexible polyester films (0.005in, DuraLar\textsuperscript{TM}, Grafix Plastics) and conductive threads (235-34 4ply HCB, V Technical Textiles Inc.) as an example. It is worth noting that the conductive thread is braided from 4 multifilament plies and each filament is a silver-coated Nylon wire. The resulting SCP actuator is formed by twisting conductive threads until they form coils and then annealing the coils thermally. The detailed process is presented in Yip and Niemeyer\cite{Yip2015}. Geometrically, the SCP actuator resembles a spring.

\subsection{Out-of-Plane Folding}
Out-of-plane (or 3D) structures are essential components of origami robots for actuation, locomotion, as well as supporting and loading functions. To build complicated 3D geometries, current origami (or Kirigami) methods require either extremely complicated 2D patterns or special multilayer laminates (and associated complex fabrication processes\cite{Felton644}), which seriously limit the accessibility of robotics. Here, we propose a new out-of-plane folding method to create complex 3D architectures from single-layer planar sheets, inspired by a very old fabrication approach--weaving. 

\begin{figure}[t]
  \center
  \includegraphics[trim=0cm 7.9cm 0cm 0.6cm, clip=true,width=0.5\columnwidth]{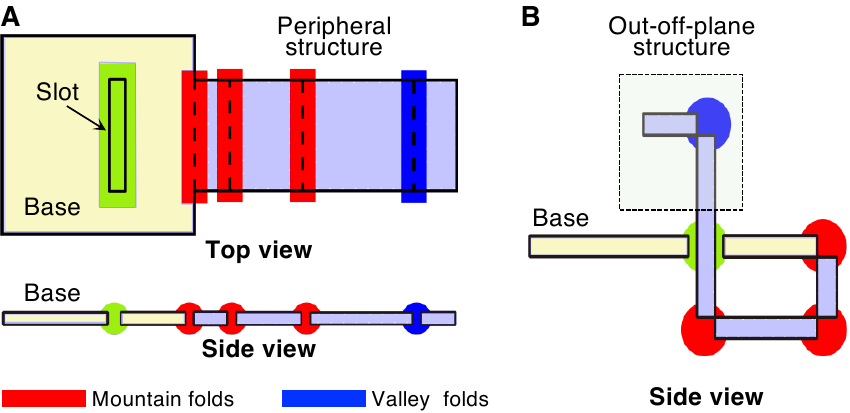}
  \caption{A schematic of out-of-plane structures formation. The red regions represent mountain folds and the blue regions are valley folds. The green regions indicate slots on the base. For visualization, no connector is shown here. (A) A patterned sheet in planar form. The figure at the top shows its top view and the figure at the bottom shows the side view. (B) The assembled 3D device with an out-of-plane structure. Usually, a fastening method is used to fix the out-of-plane structure at the intersection area (e.g. slot neighborhood).}
  \label{fig:schematicOfOOP}
\end{figure}

Essentially, the out-of-plane structures are created in a manner similar to weaving, as shown in Fig. \ref{fig:schematicOfOOP}; the out-of-plane structure is formed by making the strip-like peripheral structure go through the slot from the backside of the base. In the simplest case, there are two main components, namely a base (with a slot) and a peripheral structure (Fig. \ref{fig:schematicOfOOP}A). The base functions as a foundation to support the out-of-plane structure. The peripheral structure contains several folding lines that enable it to go through the slot on the base when properly folded. The fabrication process includes three steps: (i) 2D patterning, (ii) go-through folding, and (iii) fastening (discussed in the next section). In particular, our method can apply to any single-layer sheet material. The 2D pattern can be created with laser cutters, paper cutters, or rulers and knives. Once this patterning process is completed, we can proceed the fabrication by folding the peripheral structure in the way presented in Fig. \ref{fig:schematicOfOOP}B and guiding the structure to go through the corresponding slot from the backside of the base structure. Since sheet materials might be flexible, we use turn-up structures to stiffen the base and improve the rigidity of the resulting systems (Fig. \ref{fig:support} and \ref{fig:bistableBeam}). Lastly, established fastening methods, such as those realized with glue and tab-slots\cite{Onal_2015}, could be applied at the intersection between the base and peripheral structure to finalize the fabrication. Here, we demonstrate this method by building 3D support structures and bistable beams.

\subsubsection{\textbf{Support Structure}}
\begin{figure}[t]
  \center
  \includegraphics[trim=0cm 3cm 0cm 0.6cm, clip=true,width=1\columnwidth]{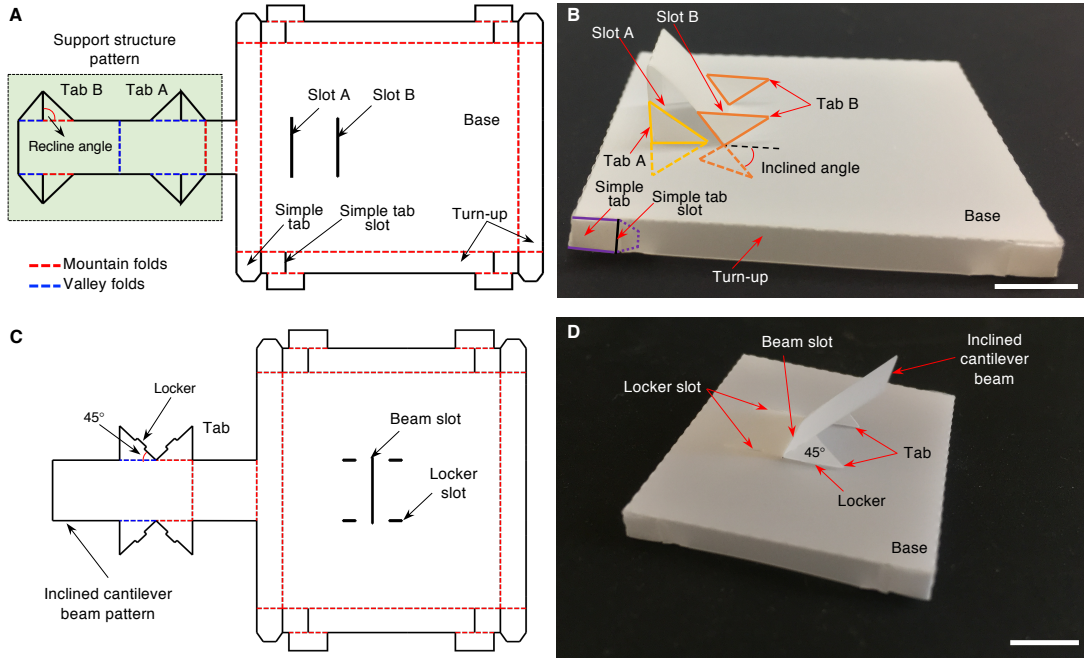}
  \caption{Fabrication of a support structure and an inclined cantilever beam.
  The red dashed lines represent mountain folds and blue lines are valley folds. Scale bars, 1 cm. (A) The 2D pattern of the support structure with a base. (B) The assembled 3D structure of the support structure, fastened by double-rib tab-slot connections on a base. The stiffness of the support structure is enhanced by forming a triangular framing. (C) The 2D pattern of the inclined cantilever beam with a base. (D) The assembled 3D structure of the beam structure, fastened by special double-rib tab-slot connections on a base. The special tab with locker matching with the corresponding slot on the base can shape the inclined angle (we use 45\textsuperscript{$\circ$} as an example here) of the cantilever beam.} 
  \label{fig:support}
\end{figure}

Fig. \ref{fig:support} illustrates the design and fabrication process of a simple out-of-plane support structure on a base. The 2D pattern of the device is shown in Fig. \ref{fig:support}A. There are two slots on the base because we need to construct a triangle frame to enhance the stability of the support structure. Two types of tab-slots are used to fasten the structure (details will be elaborated in the next section). Firstly, we implement the 2D fabrication with a cutting machine (Silhouette CAMEO 2, Silhouette America, Inc.), which is relatively inexpensive and has good resolution. Secondly, we fold the patterned 2D structure of the support structure downward and make it go through Slot A. The position of the support structure is fixed by Tab A at Slot A. The 2D pattern is then forced to go through Slot B from the frontside of the base and fixed by Tab B at Slot B. However, the base of the support structure is not rigid or strong enough due to the intrinsic flexibility of the materials. We add the turn-up structures (connected by simple tabs, Fig. \ref{fig:support}B) along the edges of the base to strengthen it. This support structure can sustain large loading, despite the low rigidity of the material. Other similar supporting or framing structures, such as cantilever beams, can be created with the same method. So far, we have demonstrated the ability of our method to build isolated 3D structures, such the support structure we presented. We are also capable of creating more complicated structures and mechanisms, which we will manifest in the following section with a bistable buckled beam.

\subsubsection{\textbf{Bistable Beam}}
\begin{figure}[t]
  \center
  \includegraphics[trim=0cm 0.5cm 0cm 0.6cm, clip=true,width=0.5\columnwidth]{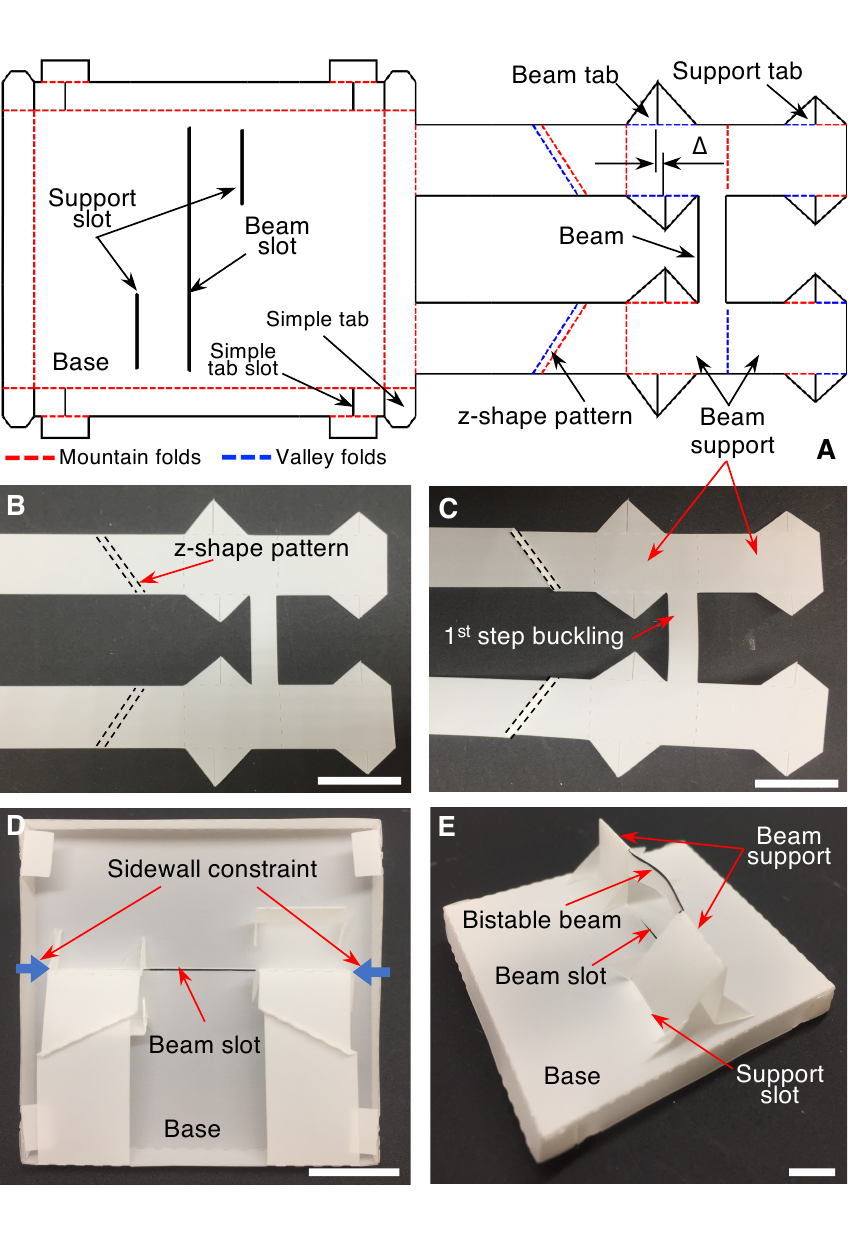}
  \caption{The fabrication of a bistable beam. The red dashed lines represent mountain folds and blue lines are valley folds. Scale bars, 1 cm. (A) The 2D pattern of the bistable beam with a base. (B) A high-resolution photo of the z-shape pattern. (C) A folded z-shape pattern causing first-step buckling of the beam. (D) The beam is further buckled by the geometry constraint from the sidewalls of the beam slot. The turn-up structure is removed for visualization. (E) A high-resolution photo of the bistable beam.}
  \label{fig:bistableBeam}
\end{figure}

Based on the method we proposed, we can create a bistable buckled beam, as shown in Fig. \ref{fig:bistableBeam}. In order to make the desired bistable beam out of a planar sheet, we need to create a compliant beam with rigid supports that can apply axial compression on the beam and result in buckling. With this concern, the major fabrication process is similar to the aforementioned standard procedure, while there are several problems that need to be addressed. To generate the axial compression, we developed an approach with two steps: (i) z-shape folding and (ii) geometry confinement. The z-shape folding pattern (Fig. \ref{fig:bistableBeam}A and B) results in a first-step compression on the beam by decreasing its axial span (Fig. \ref{fig:bistableBeam}C). Then we feed the beam through the beam slot on the base. The side walls of this slot further constraint the beam in axial direction (Fig. \ref{fig:bistableBeam}D), resulting in the desired bistable beam. Finally, the supports of the beam are fixed on the support slots to immobilize its boundary condition. This bistable beam has two stable equilibrium states and can transform a continuous actuation into a sequence of discrete, transient motions, which can be potentially used for actuation\cite{chen_harnessing_2018}, switching\cite{rothemund_soft_2018}, energy harvesting\cite{shan_multistable_2015}, and deployment mechanisms\cite{chen_integrated_2017}. 

It is worth noting that the bistable beam slightly bends upward owing to the asymmetric confinement (no confinement on the top of the beam supports) from side walls. To compensate for this asymmetry, we intentionally differentiate the heights of the fastening tabs with a discrepancy $\Delta$, as shown in Fig. \ref{fig:bistableBeam}A. This discrepancy leads to a downward bending of the beam. Combined with the initial bending, it moderately improves the symmetry of the bistable beam while avoiding complicating the system by adding accessories. For better symmetry, another geometrical confinement on the top of the frames of the beam could be applied. 

Moreover, our folding strategy can be harnessed to build more complicated structures. For example, we can create a 3D device with multiple out-of-plane structures, which we present in Section \ref{sec:example}. Additionally, we can create hierarchical 3D structures by employing our weaving-inspired out-of-plane folding strategy.
Thanks to its low cost and simplicity, our method has various potential applications. Otherwise, we can use special multilayer laminates to construct 3D architectures\cite{Whitney_2011}. However, this method requires specific materials, complex fabrication processes and associated expensive equipment, which seriously limits the accessibility of robotics though we can obtain more delicate and versatile 3D structures. In terms of performance, the usage of multilayer laminates may reduce the strength-to-weight ratio of resulting devices.

\subsection{Structure Fastening}
Although structure fastening can be done in many ways, here we aim at minimizing the resources. There are two types of connections we are interested in. One is the fixation between origami structures and the other is the connection between the origami structures and SCP actuators. The reason why we are concerned about the connection between origami structures and SCP actuators is that these actuators are essential components in our scheme of electromechanical systems. They function to complete the circuitry as electrical paths and to supply momentum as actuators. These two different types of connections will be addressed in following sections.

\subsubsection{\textbf{Origami-Origami Connection}}
Previously, the mechanical tab-slot mechanism\cite{Onal_2015} has been used extensively as a fastening method for origami robots and related devices. However, this kind of tab-slot joint can only ensure the position accuracy but not the angular accuracy of the connection. In addition, this tab-slot method has strict geometric requirements for the connecting structures, which precludes the successful implementation of the go-through folding. 

In this work, we propose a double-rib tab-slot mechanism to address the drawbacks of the current tab-slot connection. The basic configuration of this double-rib tab can be found in Fig. \ref{fig:support}A (e.g. Tab A and Tab B). Taking Tab A as an example, its two triangular ribs on each side function together as a mechanical clamp on the corresponding slot, fixing the position of the support structure. The ribs can be folded into $90\textsuperscript{$\circ$}$ against its own foundations and strengthen the structure to make it stand vertically against the base (Fig. \ref{fig:support}B); in other words, this tab helps to shape the geometry of the support structure. More specifically, the ribs are folded by 180\textsuperscript{$\circ$} against their foundations in order to go through the narrow slots, and then unfolded to $90\textsuperscript{$\circ$}$ to force their foundations to stand vertically against the base. We can also design the recline angle $\theta$ (Fig. \ref{fig:support}A) to a value consistent with the inclined angle $\phi$ (Fig. \ref{fig:support}B) to help shape its geometry. For example, we can build a 45\textsuperscript{$\circ$}-inclined cantilever beam via the following steps. We set the recline angle of the tab as 45\textsuperscript{$\circ$} as shown in Fig. \ref{fig:support}C. The match between tab lockers and corresponding locker slots forces the tab to stay at 90\textsuperscript{$\circ$} against their foundation, which guarantees the desired incline angle for the the cantilever beam at 45\textsuperscript{$\circ$}, shown in Fig. \ref{fig:support}D. It is worth noting that it is not necessary to add such locker-slot mechanisms when the targeted inclined angle of the out-of-plane structure is 90\textsuperscript{$\circ$}. In particular, for the bistable beam formation, this approach not only helps to straighten the support of the beam, but also greatly increases the stiffness of the beam's boundaries (Fig. \ref{fig:bistableBeam}E). The enhanced stiffness is also crucial for improving the symmetry of the bistable beam. In addition, the widths and lengths of the slots are intentionally designed to be small enough to better position the out-of-plane structure (Fig. \ref{fig:schematicOfOOP}B) while they should be sufficiently large to enable assembly. Empirically, the width should be set to around 1.05 times of the thickness of the sheet material while the length should be same as the width of the foundation in the support structure pattern.

Nevertheless, this double-rib tab seems overqualified when the featured size of the connection becomes small and the typical loading of the connection is relatively small. We instead use simple tab-slot connection (Fig. \ref{fig:support} and Fig. \ref{fig:bistableBeam}). The simple tab has a tapered tip to facilitate its insertion into the slot and the friction between the tab and the slot makes the tab stay attached. Thanks to the simple structure and assembly process of this tab-slot mechanism, it is expected to have wide applications in origami-inspired fabrication.

There are other fastening methods available for origami structure connections, with the most common alternative being glue. By using glue, we can take advantage of simpler origami structures, as extra patterns for tabs and slots are no longer necessary (although connecting patterns are still required). Meanwhile, the assembly process is easier without constructing the tab-slot connections. However, it is worth noting that additional glue will increase the cost and the required resources for creation, which may prohibit the accessibility of such robots. At the same time, glue is largely irreversible, which means we are not able to disassemble the resulting devices back into planar form for compact store and transportation. On the contrary, we can disassemble the tab-slot connections with ease and thus the resulting devices. One of the most important behavioral parameters of fastening is the connection strength. Our new double-rib tab-slot mechanism utilizes rib as a mechanical stop to construct the connection, which indicates the connection strength of our mechanism largely depends on the rib's strength that is normally smaller than the tensile strength of the constitutive material. On the other hand, glue can usually form bonding that is stronger than the constitutive materials. Hence, our double-rib tab-slot mechanism is beneficial for applications in which extremely strong connections are not required or resources are limited.

\subsubsection{\textbf{Origami-Actuator Connection}}
The actuator functions similarly to shape memory alloy (SMA) actuators; it contracts when it is heated through the power applied across the actuator. Geometrically, its spring-like shape makes it difficult to be fixed, especially on origami structures (e.g. the support, Fig. \ref{fig:support}). On the other hand, the radial expansion-shrinking cycle of the actuator during operation requires the fastening method to compensate for this variation in geometry. Hence, it is worthwhile to investigate various methods to address this type of connection between actuators and origami structures, as it is common in electromechanical systems. 

\begin{figure}[t]
  \center
  \includegraphics[trim=0cm 15cm 0cm 0.6cm, clip=true,width=1\columnwidth]{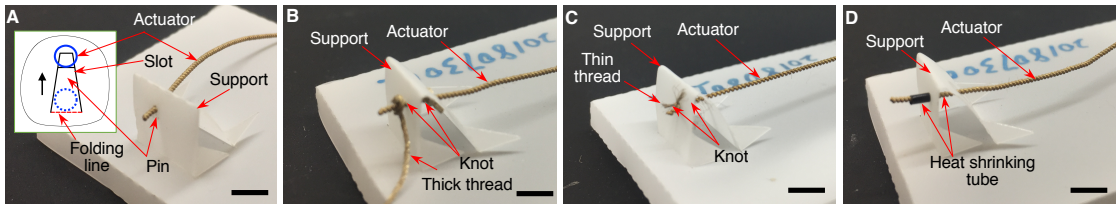}
  \caption{Connections between support structures and actuators. Scale bars, 1 cm. (A) The coils of the actuator function as mechanical stoppers to attach itself onto an origami fastener on the support structure. (B) Thick conductive threads are used to fasten the actuator on the support structure by tying knots on its both sides. (C) Thin conductive threads are employed to immobilize the actuator on the support structure through tying knots on its both sides. (D) Heat shrinking tubes are harnessed to fix the small pin on the support structure and the actuator, in order to fix the position of the actuator.}
  \label{fig:origamiActuator}
\end{figure}

To minimize the required resources, we explored several methods to solve this problem, presented in Fig. \ref{fig:origamiActuator}. Firstly, we choose not to use any additional material and to use an origami fastener instead. This fastener has a tapered slot and a pin. As shown in Fig. \ref{fig:origamiActuator}A, the pin is connected with the slot along a folding line. To assemble the device, the actuator is pulled through the slot from the wider (button) side with the pin folded. Then the actuator is pushed into the narrow (top) side from a desired position. Thanks to the spring-shape structure of the actuator, the coils themselves function as mechanical stoppers to fix the position of the actuator in its axial direction. The small pin is finally imposed back into its original position to prevent the lateral movement of the actuator. The stability and robustness of this connection mainly depends on the strength of the coil. Due to the low Young's modulus of the actuator (or conductive threads), this fastener only works in low-load applications. 

Secondly, we use threads to attach the actuator to the support structure (Fig. \ref{fig:origamiActuator}B). The same conductive thread for creating the actuator is used to form knots on both sides of the support structure, where these knots serve as mechanical stoppers to pin the actuator. This method was a success in the first few operations but failed because the knots were loose. Through careful examination, we found that the looseness resulted from insufficient friction between the threads, as their contact area was very limited. 

Naturally, as an improved version of the second method, we replace the thick thread with thin thread (unbraided from the thread) and perform the same assembly process (Fig. \ref{fig:origamiActuator}C). After several tests, the knots remained tight. The result showed that the thin thread worked much better than the previous method, indicating that thin (and softer) thread could supply much larger friction to hold the knots tight. 

Lastly, if we do not particularly aim for simple structures and minimal resources, we can replace the thin thread with a heat shrinking tube. The advantages in using this tube include: (i) the convenience in assembling and (ii) the high robustness of connection. When the actuator is heated, its radius increases while the tube shrinks, resulting in a tighter contact at the interface between the tube and the actuator, which avoids detachment or clearance creation.

In summary, the most challenging issue while forming a solid connection between a SCP actuator and an origami structure is caused by the the actuator's thermal expansion, which can initiate a separation between the actuator and origami structure resulting in a connection failure. In this section, we address this issue by exploring three different methods to compensate for the thermal expansion. The first method requires minimal resources, although this method has limited applications due to the low loading ability resulting from the intrinsically unstable coil structure of the SCP actuator. The second method only needs threads for fastening. Nonetheless, it is rather challenging to tie knots at specific positions on the spring-shape actuator. The last approach requires extra heat shrinking tubes and a source of heat. In this method, the resulting connection could be very strong and robust. In addition, the assembly process can be much easier than that of the second method. Nevertheless, the requirement for extra materials and tools can increase the cost and reduce the accessibility of resulting devices. In conclusion, the first method does not require extra resources but creates the weakest connection of the three. The second method has a rather good connection strength and requires few additional resources. Although the last method can form the strongest connections, it needs extra heat shrinking tubes and a heater that are relatively expensive. In this paper, we use the second method in order to minimize required resources while obtaining modestly strong connections. For other purposes, designers may choose their options based on their desired connection strength over cost and required resources.

\subsection{Electrical Contact Forming}
\label{sec:conductivity}
\begin{figure}[t]
  \center
  \includegraphics[trim=0cm 1.7cm 0cm 0.6cm, clip=true,width=\columnwidth]{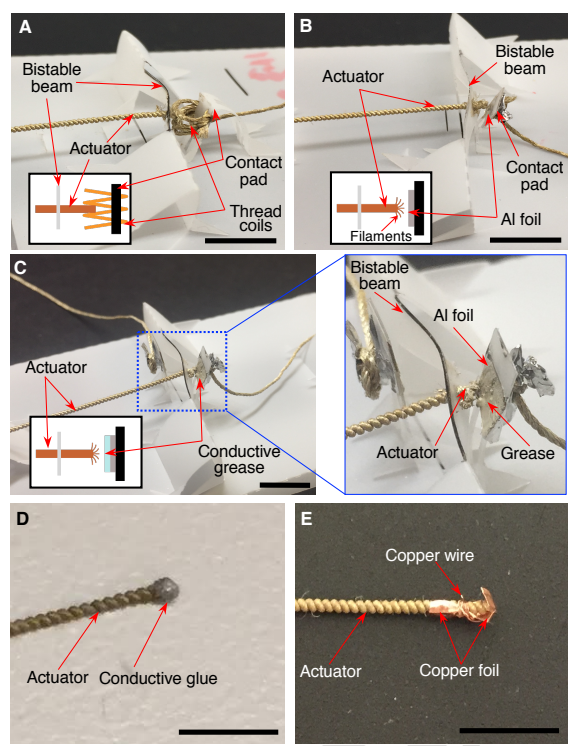}
  \caption{Electrical contact of an actuator. Scale bars, 1 cm. (A) The conductive thread is weaved into coils to form an actuator-conductive thread contact. (B) An aluminum foil forms the other terminal to create an actuator-Al foil contact. (C) Based on the above design, a small volume of conductive grease is added to the contact interface to enhance its conductivity. (D) Conductive glue is applied at the end of the actuator to form a new conductive terminal, resulting in a new conductive glue-copper foil contact. (E) Another copper foil is harnessed to electrically connect with the actuator; thus, a new copper foil-copper foil contact is built.}
  \label{fig:actuatorContact}
\end{figure}

Since the SCP actuator can function as both the actuator and the wire, it is necessary to consider the electrical connection formed between the actuator and other conductive agents (e.g. conductive thread). There are two different kinds of electrical contacts, stationary contact and dynamic contact (similar to a switch). The stationary contact could be formed easily through knotting\cite{hansora_performance_2015} (with conductive thread or copper wire) while the dynamic contact remains challenging due to the special configuration of the actuator. In this section, we investigate methods to form a robust dynamic electrical contact with the actuator as one of its terminals. Aiming at minimizing resources, various potential approaches are displayed in Fig. \ref{fig:actuatorContact}. The bistable beam-actuator-conductive agent structure is used as an example. Specifically, one end of the actuator is fixed on the bistable beam with its small overhanging segment as one terminal of the contact. A (flexible) pad is placed on the corresponding position to form the other terminal of the contact. 

\subsubsection{\textbf{Actuator-Thread}}
Firstly, we choose to use the cylindrical surface of the actuator as the contact terminal. The idea is to use conductive thread as the other contact terminal and avoid using extra components. We weave the conductive thread on the pad into loose conductive coils, as shown in Fig. \ref{fig:actuatorContact}A. When the actuator is pushed toward the conductive coils, its cylindrical surface forms contact with the coils. However, the pressure between the cylindrical surface and conductive coils is not big enough to create a robust electrical contact. The only way to improve this method is to apply a lateral pressure during the contact, but it will make the contact structure overly complicated. 

\subsubsection{\textbf{Actuator-Aluminum Foil}}
Geometrically, the cylindrical surface of the actuator makes it difficult to form solid contact with other conductive agents, encouraging us to choose the cross-section of actuator's terminal as the contact terminal. Meanwhile, due to the common axial contraction/relaxation moving pattern of the actuator, the cross-section of the actuator's end is a better candidate for the contact terminal. We verify this idea with another two different configurations. 

Firstly, we construct an actuator-aluminum foil contact (Fig. \ref{fig:actuatorContact}B). The Al foil is attached to the pad. The pressure of this contact is generated by the recovery force of the bent flexible contact pad. This contact configuration can ensure enough pressure on the contact interface. To increase the contact surface, we slightly unbraid the end of the actuator to form a soft terminal. When the contact is formed, the pressure forces the soft terminal to have a tighter contact with the Al foil. Nevertheless, in our experiment, the duration of this contact was limited. We examined the soft terminal of the actuator after this failure. We found that the conductive filaments on the terminal were melt due to the massive heat during actuation. This melting destroyed the conductivity of the thread and eventually led to the loss of the contact. 

Naturally, our second configuration refines the actuator-Al foil contact through protecting the conductive filaments from melting. Hence, conductive grease (Volume electrical resistant, 0.01 ohm/cm) is added onto the interface between the terminal of the actuator and the Al foil (Fig. \ref{fig:actuatorContact}C). This grease, on one hand, prevents the melting of conductive filaments by facilitating heat dissipation. On the other hand, thanks to its liquid state, the grease increases the contact area by a huge amount. This contact has various advantages, including simplicity, low cost, long lifetime, etc., and these advantages are sufficient for most situations within our design scheme. In the future, we will explore ways to eliminate the conductive grease, in order to reduce the required resources.

\subsubsection{\textbf{Other Electrical Contacts}}
Without considering cost and accessibility, we explore two more methods to form reliable electrical contact. Essentially, we make an effort to convert the actuator-agent direct contact into an agent-agent format. This agent-agent contact is realized by introducing an intermediate conductive agent to form a stationary connection with the actuator beforehand. Firstly, as shown in Fig. \ref{fig:actuatorContact}D, we use conductive glue to form a contact terminal on the actuator and, in turn, use this new contact terminal to form contact with the copper foil (which has better oxidation resistance than Al foil). This type of contact has high robustness and good conductivity, but it requires much larger pressure due to the small contact area, which limits its application range. Secondly, as shown in Fig. \ref{fig:actuatorContact}E, we connect the actuator with a copper wire and use this copper wire to form contact with a copper foil. Thus, we have created a copper-copper interface, which results in a robust and low-resistance electrical contact. The reason why we use copper wire as the intermediate agent is that the small copper wire can fit into the interval between the coils of the actuator to create a solid electrical connection.

In general, the performance, especially the conductivity, of dynamic electrical contacts with SCP actuator majorly depends on the interface area and pressure. Due to its intrinsic thin spring shape, here we need to seek ways to increase the contact area and thus the contact conductivity. Five methods have been investigated and can be summarized as follows. 1) Actuator-thread contact is the most resource-efficient method since we only need threads without any extra materials. However, we have to carefully design the contact configuration that can introduce high pressure to ensure the conductivity. This requirement can overly complicate the design and assembly process, though it may be a potential candidate for applications where very limited resources are supplied. 2) Actuator-aluminum foil contact has a relatively simple structure and thus an easy assembly process. Nevertheless, to obtain high conductivity, a high pressure at the contact is still required as the first method. Also, the conductivity of the contact can be fairly sensitive to the working temperature, which makes it only applicable to low-temperature applications. 3) Actuator-conductive grease-aluminum contact is a refined version of the second method by applying conductive grease to mediate the low conductivity and overheating issues. However, we need extra conductive grease, which will raise the cost. Meanwhile, the pasty form of the intermediate can cause varying resistance of the contact, which may affect the performance of resulting systems. 4) (Actuator-)conductive glue-copper foil contact has good conductivity and high robustness. However, it needs large pressure and extra conductive glue, which limits its application range. 5) The last one is (actuator-)copper-copper contact. Despite the challenging assembly process due to its small feature, this method can ensure a robust and low-resistance electrical contact thanks to the direct copper-copper interface. This method also requires very thin copper wires and copper foils, which is not applicable for resource-limited applications. To sum up, contacts of methods 1), 2), and 4) have the lowest conductivity due to their small contact area. The last type of contact has the highest conductivity though it requires extra expensive materials. In this paper, we temporally choose to use the actuator-conductive grease-aluminum contact, the third method, with aims to realize a balance between conductivity and cost. Other designers are allowed to select different methods according to their own requirements for the conductivity of the contact over cost. In future, more low-cost and easy-to-assemble methods will be explored.

\subsection{Thermal Budgeting}
\label{sec:heat}
The SCP actuators are essential components in our design space. Due to the special thermal (Joule heating) actuation, the thermal budgeting of the actuator can be crucial to the functionality and stability of the resulting systems. Primarily, we focus on the temperature control of the resulting systems to ensure their functionality under the interference of temperature variation. Here, two categories of approaches, including passive and active cooling, have been investigated. To demonstrate our method, we use the actuator-Al foil contact with conductive grease (Fig. \ref{sec:conductivity}C) as an example. At room temperature (24\textsuperscript{$\circ$}), the device is placed on a testing platform, and a thermal camera (Compact, Seek Thermal, Inc.) is placed above the device.

\begin{figure}[t]
  \center
  \includegraphics[trim=0cm 8.7cm 0cm 0.6cm, clip=true,width=0.5\columnwidth]{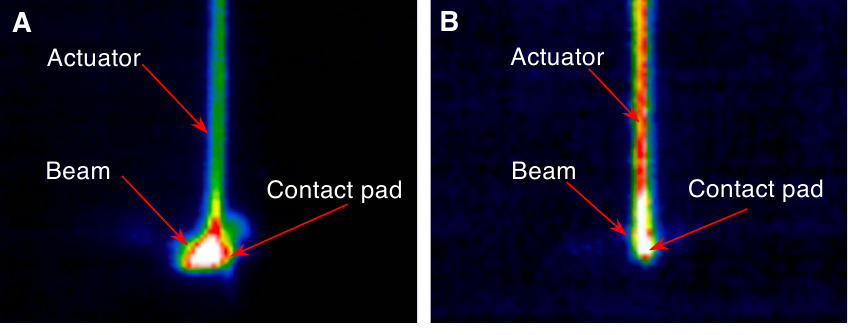}
  \caption{Temperature distributions of the contact under different circumstances. The thermal images of the contact during operation in standing air (A) and in forced air to actively control the temperature of the system (B).}
  \label{fig:temperatureImage}
\end{figure}

\subsubsection{\textbf{Passive Cooling}}
We first use standing air as the cooling source. Once the power (constant current, 0.45 A) is supplied, the actuator starts to heat up and contract. Meanwhile, the temperature of the beam and flexible contact pad also increase (Fig. \ref{fig:temperatureImage}A) due to the local heat zone. After about 4s, the power turns off. We find that the beam and flexible contact pad are thermally deformed even when the working temperature is below the polyester's glass transition temperature $T\textsubscript{g}$ (approximately 60\textsuperscript{$\circ$}). The SCP actuator usually works at a temperature close to its own glass transition temperature (approximately 60\textsuperscript{$\circ$})\cite{forman_modifiber:_2019}, which is higher than that of polyester. Thus, the standing air cannot guarantee the functionality of the contact in the current configuration. Though the standing air cannot prevent the polyester structures from melting, a media with higher thermal conductivity coefficient, such as water, should be able to avoid the melting. Thus in an experiment, we immersed the device into the water. Under the same conditions, the actuator barely contracted even under a current of 4.0 A. This method limits our devices to aqueous applications. Still, it is possible to enable the operation with a liquid media that has a suitable thermal conductivity coefficient; nevertheless, as this requires a lot of resources, we have not gone into further research in this direction.

\subsubsection{\textbf{Active Cooling}}
We then use forced air to enhance the cooling process and to protect the polyester components. The result, shown in Fig. \ref{fig:temperatureImage}B, indicates that the beam and flexible contact pad have relatively low temperature compared to in standing air. It is worth noting that this method consumes more energy due to the heat dissipation through enhanced thermal convection. 

Again, if we are not very concerned with minimizing required resources, we can address thermal budgeting in several other ways. For example, we can replace the polyester with high thermal-resistance materials, such as polyether ether ketone (PEEK, $T\textsubscript{g}=143\textsuperscript{$\circ$}$). These high thermal-resistance materials can sustain high working temperature above the glass transition temperature of the SCP actuator, which can eliminate the overheating issue. Nevertheless, these material are far more expensive than polyester and will increase the cost and reduce the accessibility of resulting systems. On the other hand, by harnessing active cooling, not only do we protect the origami components, but also have more control on the cooling speed of the system. This added control may enable more sophisticated performances of the resulting devices, such as time-varying behaviors by dynamically adjusting cooling. In conclusion, the active cooling allows higher working temperature even with low heat-resistant materials at the cost of extra cooling systems. Otherwise, we can use expensive, high heat-resistant materials at high-temperature environment or passive cooling at low-temperature applications.

\section{Electromechanical Oscillator}
\label{sec:example}
In this section, we demonstrate the feasibility of our new strategy with a functional electromechanical device, an electromechanical oscillator. This oscillator is an analogy to a relaxation oscillator. Under a constant-current power source, our electromechanical oscillator is capable of inducing current oscillation without an electronic circuit. We first present the design and fabrication process of the oscillator, and then validate its functionality through experiments and testing.

\begin{figure}[t]
  \center
  \includegraphics[trim=0cm 0.5cm 0cm 0.6cm, clip=true,width=0.5\columnwidth]{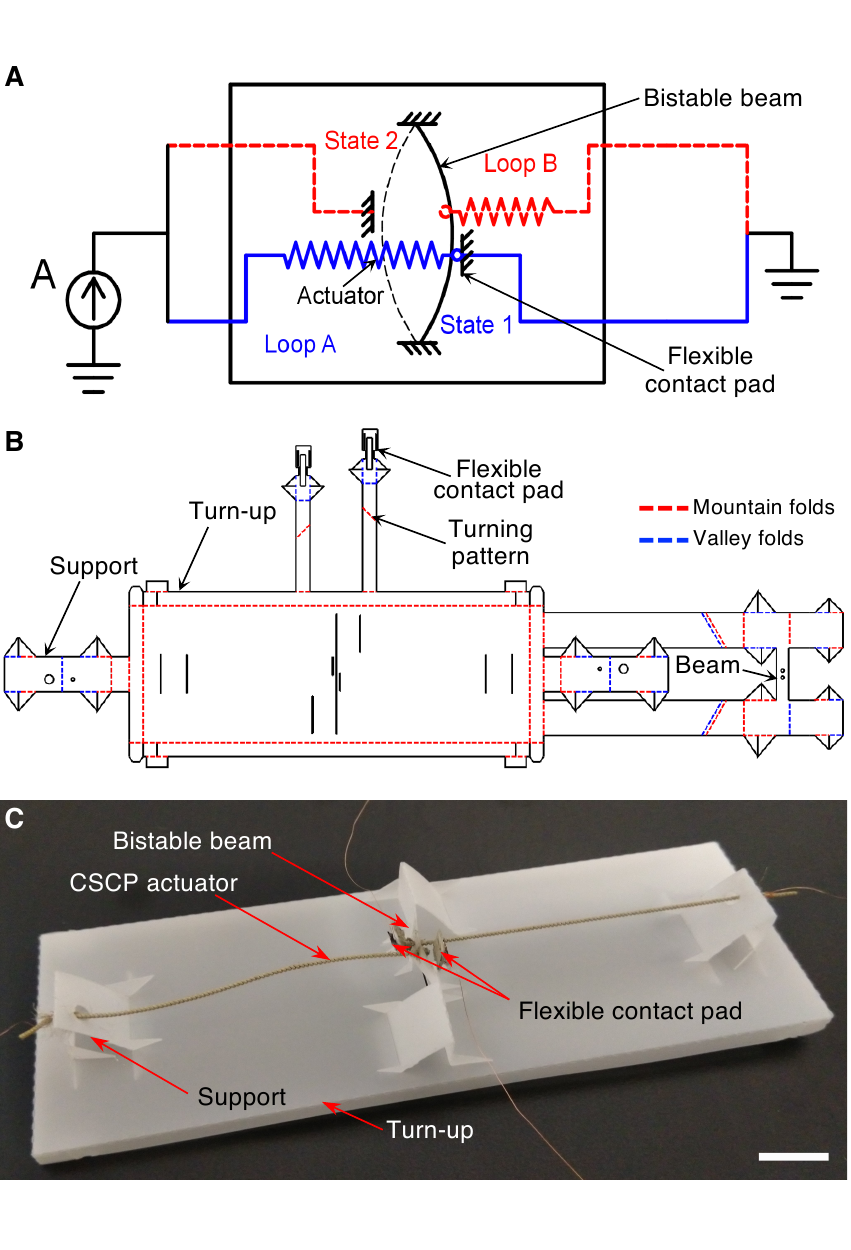}
  \caption{An electromechanical oscillator. Scale bars, 1 cm. (A) The schematic and mechanism of the foldable oscillator. (B) 2D pattern of the origami structure of the foldable oscillator. The red dashed lines represent mountain folds and the blue lines are valley folds. (C) The assembly of the foldable oscillator.}
  \label{fig:MLDesignAndfabrication}
\end{figure}

\subsection{Design and Fabrication}
The mechanism of this electromechanical oscillator is shown in Fig. \ref{fig:MLDesignAndfabrication}A. The bistable mechanism functions as a double-pole, single-throw switch for both Loop A and Loop B in opposite phases. The actuators serve two functions: one function is to complete the circuitry, while the other is to drive the snap-through motion of the bistable mechanism. Under this circumstance, once power is supplied, the electromechanical oscillator starts to oscillate: (1) initially, Loop A is closed. The corresponding actuator begins to heat up and gradually contract along its axial direction; (2) upon reaching its critical displacement, the bistable mechanism snaps to the other stable position; Loop A is now open while Loop B is closed; (3) Actuator A begins to cool down while Actuator B begins to heat up. The previous steps then execute for the opposite loops; after this, the entire actuation process is completed. As the system repeats this entire process periodically, the electromechanical oscillator behaves similarly to an electrical oscillator. 

To realize the electromechanical oscillator as shown in Fig. \ref{fig:MLDesignAndfabrication}A, we need to construct the frames for attaching actuators, a bistable beam for loop switching, and flexible pads for electrical contacts. These complicated 3D structures are very challenging to be generated by using conventional folding strategy while can be easily created through our weaving-inspired out-of-plane folding and structure fastening principles as shown in Fig. \ref{fig:MLDesignAndfabrication}B. In addition, the actuators need to be stably fixed on support structures and the bistable beam. Due to the thermal expansion of SCP actuators, the fastening would be very challenging. However, this issue can be solved by using our methods that can compensate the expansion to ensure solid connections. The electrical contacts between actuators and flexible pads can be implemented by our proposed actuator-conductive grease-aluminum foil design. However, to ensure a solid contact before the bistable beam snap-though and an instant detachment afterwards, we harness a flexible contact pad functioning as a biased spring to generate enough pressure at the contact and thus guarantee the conductivity (see Fig. \ref{fig:MLDesignAndfabrication}C). This flexible contact pad is designed by following our out-of-plane folding principle. We use active cooling to lower the working temperature of the oscillator and control the oscillation frequency by altering the heating and cooling processes of actuators as well. In conclusion, our approach not only allows for integrated and monolithic design and rapid fabrication, but also leads to accessible, low cost, and potentially disposable designs.

The electromechanical oscillator is mainly composed of a polyester flat sheet, conductive threads, aluminum foil, and conductive grease. With SCP actuators already prepared, the oscillator is fabricated in two steps: (i) 3D origami structure formation by folding, and (ii) electromechanical assembly.

The first step is to build the 3D origami structure, including flexible contact pads, rigid supports, a bistable beam and its supports. The 2D pattern of the origami structure is shown in Fig. \ref{fig:MLDesignAndfabrication}B. The detailed fabrication processes for the support structure and bistable beam are basically the same as in Fig. \ref{fig:support} and \ref{fig:bistableBeam}. A 45\textsuperscript{$\circ$} turning folding pattern is located on each flexible contact pad, so after folding the contact pad along this pattern, it is reoriented in the direction that allows it to form a contact pair with the actuator. In order to maintain the electrical connection in the loops before the bistable beam snaps through and to instantly disconnect the loops immediately after the snap-through, we design the flexible contact pad as a cantilever structure, as shown in Fig. \ref{fig:MLDesignAndfabrication}B.

The next step is to assemble the oscillator. One end of the actuator is attached to the bistable beam through a hole, with a 1 mm-long segment overhanging (the same structure as shown in Fig. \ref{fig:actuatorContact}C). In the same manner, the other end of the actuator is fixed on the corresponding support structure (the same structure as shown in Fig. \ref{fig:origamiActuator}C) without pretension, which means there is no force pulling the bistable beam until the actuator is heated. Then a piece of aluminum foil is attached to the flexible contact pad to make the pad conductive. A small volume of conductive grease (Volume Electrical Resistant: 0.01 ohm/cm) is also spread on the contact pad to increase its conductivity (details could be found in Fig. \ref{fig:actuatorContact}C). The other half of the electromechanical oscillator is assembled with the same method. 

The oscillator is hand-crafted. However, the proposed design and fabrication strategy can guarantee the functionality of resulting devices in a great extent. For example, our out-of-plane folding method combined with the structure fastening approach can ensure the geometry accuracy, structure stiffness and connection robustness of 3D structures. However, more future work can be done to improve the consistency and robustness of electrical contacts by eliminating conductive grease.

The oscillator is a complicated electromechanical system, which would be very challenging and time-consuming for conventional design and fabrication methods. However, it only costs ~40 cents (raw materials) and takes less than one hour to build using our design and fabrication strategy. This validates the feasibility of our method, enabling inexpensive and rapid prototyping. In addition, we can create various functional devices and robotic systems in the same manner, which would greatly improve the accessibility of robotics.

\subsection{Result and Analysis}
After being fixed on a platform, the electromechanical oscillator was powered by a constant current (0.55 A) with a laboratory power supply. The electrical connection is shown in Fig. \ref{fig:MLDesignAndfabrication}A. As discussed above, we add a forced air cooling source to control the period of the oscillation by balancing the actuation rate and the heat dissipation rate of the actuator and also to prevent the bistable beam from permanent thermal deformation.

\begin{figure*}[t]
  \center
  \includegraphics[trim=0cm 13cm 0cm 3cm, clip=true,width=1\textwidth]{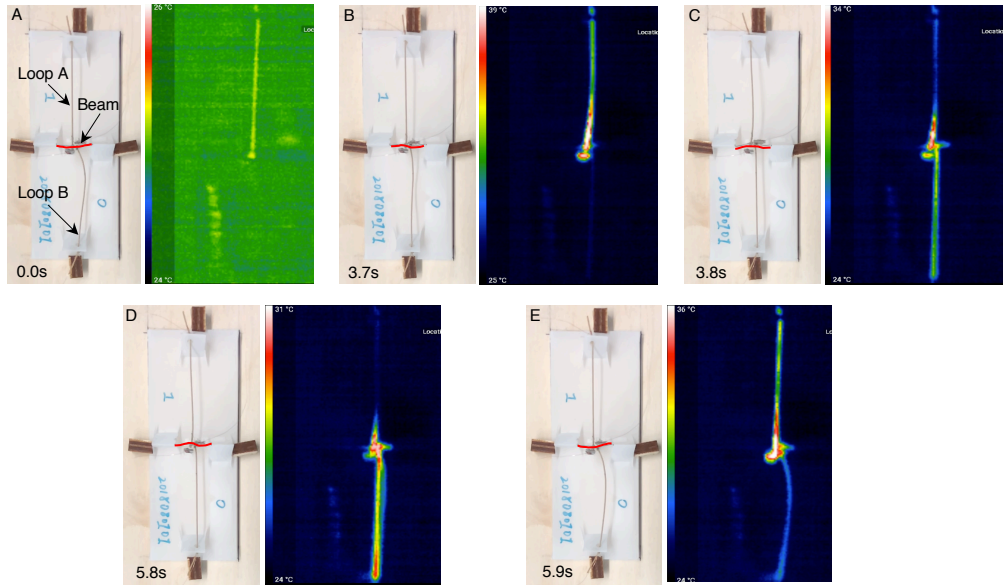}
  \caption{The first oscillation cycle of the foldable oscillator. In each figure from (A) to (E), the left side is the image of the oscillator and the right side is the corresponding thermal image obtained with a thermal camera. The red lines in the figures indicate the contours of the bistable beam in different states.}
  \label{fig:MLGraphicResult}
\end{figure*}

In our experiment, the oscillation lasted for around 20 s, during which the electromechanical oscillator went through 4.5 oscillation cycles (4 complete cycles along with an additional snap-through motion, the complete oscillation can be found in supplementary materials). We extract the result of the first cycle and present it in Fig. \ref{fig:MLGraphicResult}. Initially, Loop A was electrically connected. Under the Joule heating, the temperature of the Actuator A started to increase, leading to its axial contraction. Once the contraction of Actuator A was large enough to pull the bistable beam to its critical position (the transition mode that proceeds the snap-through, noted at about 3.7 s, as shown in Fig. \ref{fig:MLGraphicResult}B), the bistable beam snapped through and opened the Loop A at about 3.8 s (Fig. \ref{fig:MLGraphicResult}C). Meanwhile, Loop B, the loop on the opposite side, was closed. Actuator B began to heat up and contracted while Actuator A cooled down and relaxed. Transition mode of the bistable beam occured at around 5.8 s (Fig. \ref{fig:MLGraphicResult}D). At about 5.9 s, the bistable beam jumped back to its original position, completing the first oscillation. Ultimately, the electromechanical oscillator repeated the snap-through and snap-back motions, resulting in current oscillation. 

\begin{figure}[t]
  \center
  \includegraphics[trim=0cm 5.8cm 0cm 0.6cm, clip=true,width=0.5\columnwidth]{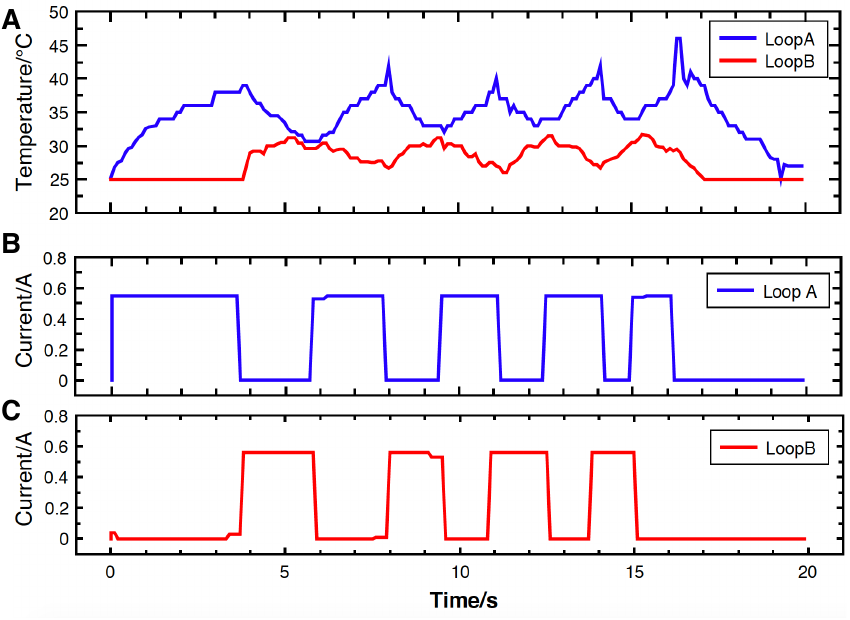}
  \caption{Curves that show the key variables (temperature and current) as a function of time. (A) The temperature curves of two actuators in both Loop A and Loop B. Plots of currents running through both Loop A (B) and Loop B (C), respectively.}
  \label{fig:MLData}
\end{figure}

To further understand the experiment, we extracted and plotted the currents running through both Actuator A and B as well as their corresponding temperatures (Fig. \ref{fig:MLData}). The current was characterized and documented by recording the readings of the laboratory power supply. In the same manner, we extracted the temperature data of the actuators from a thermal camera (Compact, Seek Thermal). As shown in Fig. \ref{fig:MLData}B and C, two loops closed and opened sequentially and asynchronously, each shifted by 180\textsuperscript{$\circ$} in phase, with an average period of about 3.75 s. This phase shift also appeared in the temperature oscillation, which confirmed the fact that our electromechanical oscillator could induce current oscillation with a constant power source. It is worth noting that there is a relatively large difference between the average temperatures of the two actuators. This difference stems from both structural and thermal asymmetries of the prototype. i) Firstly, the structural asymmetry of the bistable beam leads to the different critical displacements and thus different critical switching temperatures. This asymmetry can be minimized by designing new mechanisms to improve the fabrication accuracy of the bistable beam. ii) Secondly, the asymmetric resistances at the two contacts account for this discrepancy as well (see Fig. \ref{fig:MLGraphicResult}). This discrepancy can be largely resolved by adopting more robust electrical contacts. For example, we can use the methods as shown in Fig. \ref{fig:actuatorContact}D and E when the cost is not the primary concern. In addition, the electromechanical oscillator (not including the power supply) only cost approximately 40 cents. In summary, this functional, low-cost foldable oscillator demonstrates the viability of our new design and fabrication strategy for building electromechanical systems in a rapid and inexpensive manner. 

After the testing, we examined the electromechanical oscillator carefully to analyze the reason that caused the oscillation to stop after 4.5 cycles. Though the thermal asymmetry of the two actuators resulted in asymmetric oscillation, it was not the essential factor that limited the oscillation, according to the thermal-electro-mechanical model\cite{yan2019}. We also noticed that the lengths of both actuators decreased and the beam, though it was still bistable, was thermally deformed into an ``S" shape after disassembled from the oscillator.
Referencing to \cite{forman_modifiber:_2019}, the SCP actuator behaves substantially as a one-way shape memory alloy, which means it needs resetting to achieve reversible behavior. Accordingly, a theory was proposed: the length of the actuator decreases over time, leading to a decreasing actuation stroke without resetting; this decrease in the actuation stroke, in turn, results in the failure of the actuator to drive the bistable beam snap through. Hence, the oscillation stops. However, the actuator is still powered, which causes an unusual temperature rise of the actuator, especially at the region where the actuator contacts with the bistable beam (see Fig. \ref{fig:MLGraphicResult} and supplementary material). This unusual temperature rise results in the thermal deformation of the bistable beam. To prolong the oscillation, we can find a new type of two-way actuator or explore methods to stabilize the SCP actuator. For the latter, reference \cite{forman_modifiber:_2019} shows a promising method to stabilize the SCP actuator. In addition, we can reduce the resistance of the contact between the actuator and aluminum foil, which leads to lowering the working temperature of the bistable beam to avoid overheating. Further investigations will be done in near future to improve the performance of the electromechanical oscillator. On the other hand, we have built an analytical model for the oscillator \cite{yan2019,yan_analytical_2019}, which can be used to predict its dynamics and thus to optimize its design. By minimizing the maximal working temperature of the actuator, we can eliminate the overheating issue to avoid failure.

We have ended up with this design of the oscillation for its rapid and inexpensive prototyping, while it is totally possible to achieve the same functionality with other materials or fabrication methods. For example, we can use commercially available conductive actuators, such as shape memory alloy, to replace the SCP actuators. Moreover, we can harness 3D printing techniques to build the mechanical system and utilize the printed circuit board (PCB) fabrication method to construct the circuitry. We can even use heat-resistant sheet materials (e.g. PEEK) to substitute the low-cost polyester sheets to improve the thermal stability of the electromechanical oscillator. 

In this section, we harness the proposed design and fabrication strategy to implement an electromechanical oscillator. Our strategy is generalizable and versatile, capable of constructing complex 3D origami architectures, forming connections between different features, creating electrical contacts, and budgeting thermal flow. For example, our out-of-plane folding can be used to build an aerodynamic surface \cite{Rauf2020}. Similar to \cite{kim_origami-inspired_2018}, the structure fastening approach can be applied to create high structural stiffness mechanisms with the ribs functioning as lockers. The contact forming method is generally applicable for constructing electrical connections with thin conductive actuators, e.g NiTi coil actuators \cite{rus2013}. Lastly, the thermal budgeting strategy can be very useful for applications with low heat-resistant constitutive materials but extensive heat flow. Thus, when it comes to implementing the oscillator, we know how to create support structures or the bistable beam by using our out-of-plane folding method and structure fastening strategy. Based on the electrical contact forming approach, we are able to generate flexible contacts. With our design and fabrication strategy, we are capable of creating such a foldable oscillator by only ordinary single-layer sheets and conductive sewing threads, which is extremely difficult for other methods. On the basis of this strategy, we can presumably create endless functional electromechanical systems and thus robots. For example, we can build a fully foldable robot by using this oscillator as a simple control to sequencing two sets of SCP actuators to locomote.

\section{Conclusion and Future Work}
\label{sec:conclusionAndDiscussion}
In this paper, we explored the possibilities of minimizing the resources required for robot creation. In particular, we investigated a design and fabrication strategy for electromechanical systems for robots under the constraint of only using ordinary single-layer materials and conductive threads. This strategy addresses four major challenges: out-of-plane folding, structure fastening, electrical contact forming, and thermal budgeting. Aiming at minimizing the resources required for robot creation, we probed and presented various approaches to address each challenge, which resulted in several generalizable design and fabrication principles. These generalizable principles can be used to guide the design and creation of robotic electromechanical systems only using ordinary single-layer materials and conductive threads, which we demonstrate with a list of simple components. For example, we demonstrate our out-of-plane folding strategy with a support structure and a bistable beam. Meanwhile, this weaving-inspired folding strategy can be used to create other different 3D structures, such as a cantilever beam functioning as a flexible contact pad (see Fig. \ref{fig:MLDesignAndfabrication}). The feasibility of the proposed design and fabrication strategy was demonstrated with a foldable electromechanical oscillator that is capable of generating electrical oscillatory signals with merely a constant-current power supply. In addition, the oscillator is made in a rapid and inexpensive manner, as its raw materials cost approximately 40 cents. 

Our design and fabrication strategy can also be applied to other fields. Even when the accessibility of robotics (or the cost and resources required for robot creation) is not of major concern, our strategy still proves to be a valuable methodology for generating complex functional 3D structures (e.g. bistable beams). Our exploration of and insights into SCP actuators regarding their fastening, conductive contact forming, and thermal properties can also be useful in other applications. 

So far, we have been able to build functional electromechanical devices. Toward the creation of robotics, more work needs to be done in expanding these principles to the design of functional robotic systems with actuation, control, sensing and power. For example, we expect to create an integrated robot similar to the Octobot\cite{wehner_integrated_2016}, utilizing the oscillator as a controller. In the near future, we will investigate the possibility of building one-dollar robots with the design and fabrication strategy established in this paper.

Currently, our devices are hand-crafted. It would be desirable to have a user-friendly interface that facilitates the design process, overcoming the knowledge barrier. On the other hand, the folding and assembling process can be tedious and the fabrication error can compound as the complexity of the resulting devices increases. It may be worth investigating the possibility to incorporate self-assembling techniques into our design scheme.

Given our method, it is possible to build low-cost electromechanical systems for robots, which may lower the barrier in robot creation through reducing the resources required. On our long journey of increasing the accessibility of robotics, one-dollar robots is our next milestone. We believe that our design and fabrication strategy will be the foundation of creating one-dollar robots, as well as a step toward improving the accessibility of robotics.

\section*{Acknowledgements}
The authors would like to thank Ms. Yunchen Yu and Mr. Kenny Chen for their valuable comments and help.
  
This work is supported by the National Science Foundation under grant \#1644579 and \#1752575, for which the authors express thanks.


\end{document}